\documentclass[letterpaper, 10 pt, conference]{ieeeconf}

\IEEEoverridecommandlockouts                              % This command is only needed if 
\usepackage{times}
\usepackage{epsfig}
\usepackage{graphicx}
\usepackage{amsmath}
\usepackage{amssymb}
\usepackage{svg}
\usepackage{multirow}
\graphicspath{{figure/}}
\usepackage{hyperref}
\usepackage{algorithm}
\usepackage{algpseudocode}
\DeclareMathOperator*{\argmin}{arg\,min}

\title{\LARGE \bf
Monocular Camera Localization in Prior LiDAR Maps with 2D-3D Line Correspondences %in 3D LiDAR Maps 
}

\author{Huai Yu$^{1,2}$, Weikun Zhen$^{2}$, Wen Yang$^{1}$, Ji Zhang$^{2}$ and Sebastian Scherer$^{2}$% <-this % stops a space
\thanks{$^{1}$Huai Yu and Wen Yang are with the Electronic Information School, Wuhan University,  Wuhan 430072, China {\tt\small  \{yuhuai, yangwen\}@whu.edu.cn}}%
\thanks{$^{2}$ Huai Yu, Weikun Zhen, Ji Zhang and Sebastian Scherer are with the Robotics Institute, Carnegie Mellon University,
        Pittsburgh, PA 15213, USA
        {\tt\small \{weikunz, zhangji, basti\}@andrew.cmu.edu}}%
}

\begin{document}

\maketitle
\thispagestyle{empty}
\pagestyle{empty}

%%%%%%%%%%%%%%%%%%%%%%%%%%%%%%%%%%%%%%%%%%%%%%%%%%%%%%%%%%%%%%%%%%%%%%%%%%%%%%%%
\begin{abstract}
Light-weight camera localization in existing maps is essential for vision-based navigation. Currently, visual and visual-inertial odometry (VO\&VIO) techniques are well-developed for state estimation but with inevitable accumulated drifts and pose jumps upon loop closure. To overcome these problems, we propose an efficient monocular camera localization method in prior LiDAR maps using direct 2D-3D line correspondences. To handle the appearance differences and modality gaps between LiDAR point clouds and images, geometric 3D lines are extracted offline from LiDAR maps while robust 2D lines are extracted online from video sequences. With the pose prediction from VIO, we can efficiently obtain coarse 2D-3D line correspondences. Then the camera poses and 2D-3D correspondences are iteratively optimized by minimizing the projection error of correspondences and rejecting outliers. Experimental results on the EurocMav dataset and our collected dataset demonstrate that the proposed method can efficiently estimate camera poses without accumulated drifts or pose jumps in structured environments. %The code and our collected data are available at \href{https://github.com/levenberg/2D-3D-pose-tracking}{https://github.com/levenberg/2D-3D-pose-tracking.}
\end{abstract}

%%%%%%%%%%%%%%%%%%%%%%%%%%%%%%%%%%%%%%%%%%%%%%%%%%%%%%%%%%%%%%%%%%%%%%%%%%%%%%%%
\section{Introduction}
\label{intro}
% Robot localization is the prerequisite for robot motion planning and navigation.
%  For example, the light-weight aerial vehicles localization and navigation system.
% Currently,
Accurate robot localization in urban environments is in great demand for autonomous vehicles. Since GPS localization is unstable without direct line-of-sight to the satellites, LiDAR-based localization modules are often used because of the accurate range measurements. Additionally, urban scenes have relative time-invariant geometric structures (e.g., buildings), thus one-time 3D map construction can be used for long-term localization. However, the expensive cost and heavy weight of LiDAR sensors limit its wide applications. Cameras and IMUs are low-cost, light-weight and commonly available sensors and current visual-inertial based pose estimation and mapping methods are well-developed for a variety of robot systems  \cite{mur2017orb, qin2018vins, sun2018robust}. Nevertheless, state estimation methods that use only image features are prone to failures due to lighting or texture changes in the environment. Consequently, if camera localization modules can be associated with a prior 3D map, i.e., fuse the visual information with range measurements, there will be a great potential to use these light-weight and small camera modules for accurate localization in urban environments without a LiDAR sensor.
% [Currently, summarize the related works, however, ]

\begin{figure}[htbp]

	\begin{center}
		\includegraphics[width=0.99\linewidth]{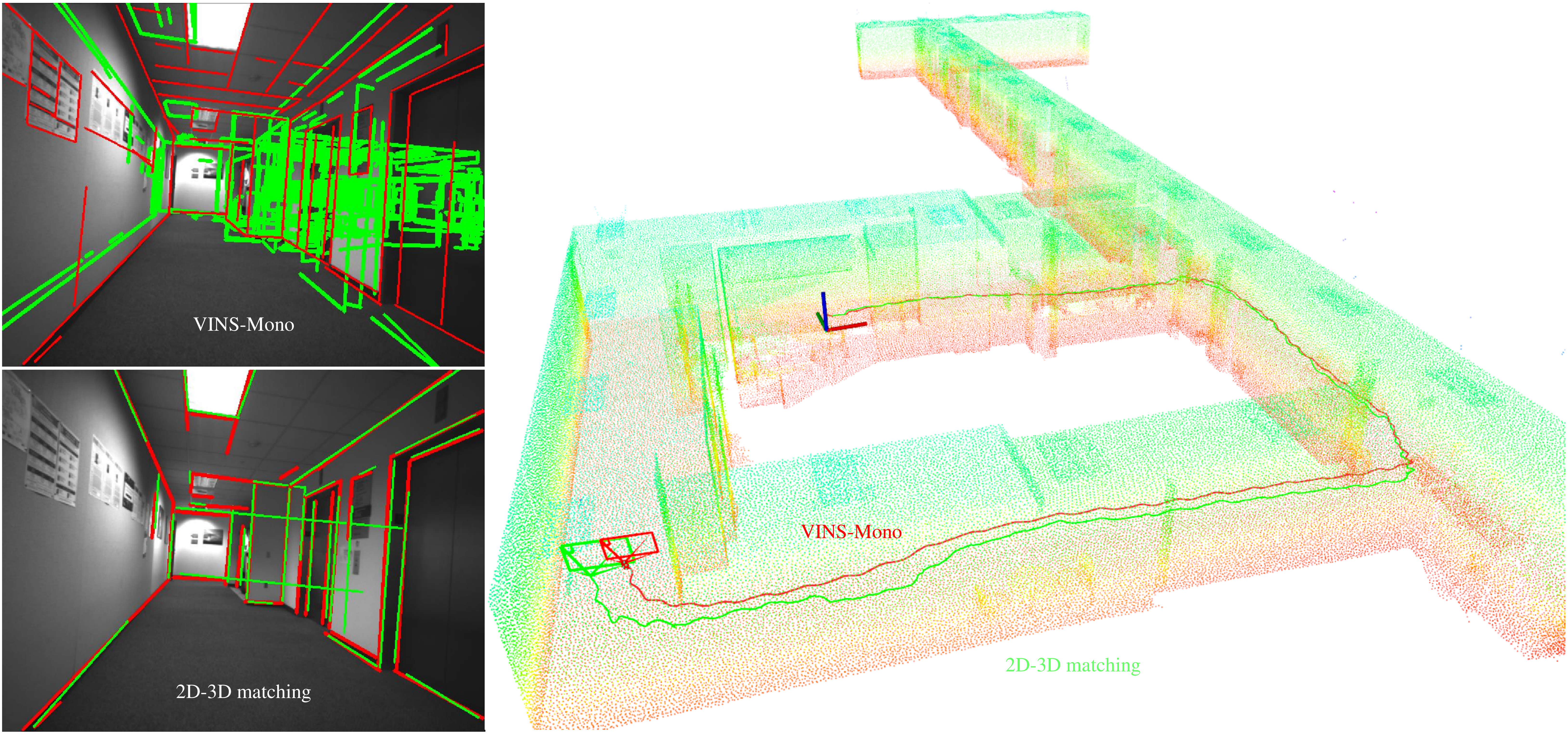}
	\end{center}
	\caption{The proposed monocular camera localization system in prior LiDAR maps of corridor. The right LiDAR map is colored by height. The red and green trajectories are the results of VINS-Mono \cite{qin2018vins} and ours, respectively. Top-left image shows the 3D line projections (green) using the estimated pose of VINS-Mono (with occlusions) and the extracted 2D lines (red), while the bottom-left image shows the 2D-3D correspondences using the pose estimation of the proposed method.}
	\label{fig:corridor}
\end{figure}
However, the fusion of image data with 3D point clouds is challenging due to appearance differences and modality gaps. Current approaches typically transfer the 3D data into 2D space or reconstruct 3D point clouds from 2D images to align data for pose estimation\cite{feng2016fast, engel2014lsd}. Based on the characteristics of urban environments, our intuition lies on the fact that the major geometric structures, such as lines and planes, can be both captured in 3D maps and 2D images regardless of appearance differences and modality gaps. The direct 2D-3D geometric co-occurrence correspondence is more robust and precise than the association of domain-transferred data. Therefore, our purpose is to directly estimate the 2D-3D geometric line correspondences for the accurate and long-term camera localization.

In this work, we propose an approach for real-time light-weight monocular camera localization in prior 3D LiDAR maps using direct 2D-3D geometric line correspondences. We assume that a coarse pose initialization is given and focus on the pose tracking in maps, which follows the related works \cite{caselitz2016monocular,zuo2019visual}. For geometric concurrent feature extraction, 3D line segments are detected offline from LiDAR maps while robust 2D line segments are extracted online from video sequences. By employing the 6-DOF pose prediction from VIO, local visible 3D lines in field-of-view (FoV) are extracted and directly matched with 2D line features to obtain coarse 2D-3D line correspondences. Finally, the camera pose and 2D-3D 
matches are iteratively optimized by minimizing the projection error of correspondences and rejecting outliers. 

The main contribution of this work is to estimate geometric 2D-3D line correspondences for camera localization, which efficiently associates every keyframe with the prior LiDAR map. The geometric line correspondences are robust to appearance changes and suitable for camera localization in urban environments. Fig. \ref{fig:corridor} shows a camera image with 2D-3D line correspondences and estimated camera poses in the LiDAR map. 

\section{Related Work}
\label{relawork}

Vision-based localization in maps is to establish correspondences between 2D and 3D modalities for improving the localization robustness and accuracy \cite{piasco2018survey}. To overcome the modality gaps and appearance differences, general approaches are using ``intermediate products" to transfer the matching into the same space, i.e., in 2D space \cite{sattler2017efficient} or in 3D space \cite{caselitz2016monocular, kim2018stereo}.

%In \cite{wolcott2014visual}, several synthetic images are generated by reflectance information of LiDAR maps and are matched with camera images by normalized mutual information. In \cite{stewart2012laps}, camera poses are estimated by minimizing the normalized information distance (NID) between the appearance of LiDAR data reprojected into overlapping images.
The first kind of camera localization in 3D maps is matching photometry in 2D image space. For most of the visual simultaneous localization and mapping (SLAM) methods \cite{mur2017orb, qin2018vins}, sparse point cloud maps are reconstructed with photometry (or visual) descriptors, then loop closure detection is conducted by matching these descriptors to decrease drifts. In image-based localization methods \cite{sattler2017efficient, feng2016fast}, large scale maps are reconstructed using SFM with visual features, these features are efficiently matched to yield 2D-3D correspondences for camera localization. Nevertheless, appearance and visual features are sensitive to illumination changes and light conditions, which make the correspondences unstable for long-term camera localization. Additionally, to handle the LiDAR maps without associated visual features, LiDAR appearance synthetic images are often used to directly match the camera images by normalized mutual information (NMI)  \cite{wolcott2014visual} or normalized information distance (NID) \cite{stewart2012laps}. Additionally, recent methods utilize colored point cloud projections \cite{pascoe2015direct} and synthesized depth images \cite{neubert2017sampling} to match with live images for camera localization. These localization methods are all trying to transfer the 2D-3D matching problem to be a 2D-2D matching problem.% which show poor performance for textureless LiDAR maps
% In  , colored point cloud maps are generated by mounted LiDAR and camera sensors, then a new camera is localized by minimizing the NID between live images and rendered images from point cloud maps. In \cite{neubert2017sampling}, depth images are synthesized from 3D LiDAR maps to match with visual images for pose estimation by minimizing mutual projections of gradient extracted from both images.  

The second strategy is matching geometry in 3D space to estimate camera poses. By using local bundle adjustment, \cite{caselitz2016monocular} proposed to match a sparse reconstructed local 3D point clouds with given 3D LiDAR maps, which solves the scale estimation problem of monocular VO system and achieves online estimation of 6-DoF camera poses. Similarly in \cite{gawel2016structure}, 3D structural descriptors are used for matching LiDAR maps with sparse visual reconstructed point clouds. In \cite{kim2018stereo, zuo2019visual}, dense local point clouds are reconstructed from a stereo camera to match to LiDAR maps, and then the matching results are loosely or tightly coupled into the VO and VIO system for optimizing camera poses. These localization methods by 3D registration obtain feasible results compared with vision-only based methods. However, the localization accuracy highly depends on local reconstruction performances. SFM suffers from scale problem and the reconstructed sparse points may not have correspondence in maps. Stereo reconstruction gives dense local point clouds but mostly is time-consuming and does not scale well for long-range depth estimation.  %SFM suffers from scale problem and the reconstructed sparse points may not have correspondences in LiDAR maps. Stereo reconstruction gives dense local point clouds but maybe time-consuming and does not scale very well for long-range depth estimation.

Compared with the local point cloud matching methods, we aim at directly extracting 2D lines from a monocular camera to match with 3D lines from LiDAR maps for camera localization, which does not rely on SFM or stereo reconstruction modules. The global 2D-3D localization is known as a kidnapped robot problem because of the feature description gap and the non-convexity \cite{campbell2018globally}.  RANSAC-based and branch-and-bound strategies are often used to maximize 2D-3D inlier correspondences for global localization without a pose prior \cite{campbell2018globally}. With the camera pose prediction from GPS or VO (VIO), we can obtain a pose prediction as the prior for local 2D-3D matching. SoftPosit \cite{david2003simultaneous} and BlindPnP \cite{moreno2008pose} are conventional simultaneous 2D-3D correspondences and pose estimation approaches with the provided pose initialization. For the camera to LiDAR sensor calibration methods \cite{levinson2013automatic, zhou2018automatic}, geometric features across images and point clouds are often matched to estimate the extrinsic transform between two sensors. Recently, some learning-based methods were proposed to directly register images to LiDAR-Map with GPS pose initialization \cite{cattaneo2019cmrnet, feng20192d3d}. These methods suffer from either high computational complexity or unstable outputs, which still need explorations for realtime camera localization in 3D LiDAR maps. 

\section{Proposed method}
\label{sec3}
\begin{figure}[htbp]
	\begin{center}
		\includegraphics[width=0.99\linewidth]{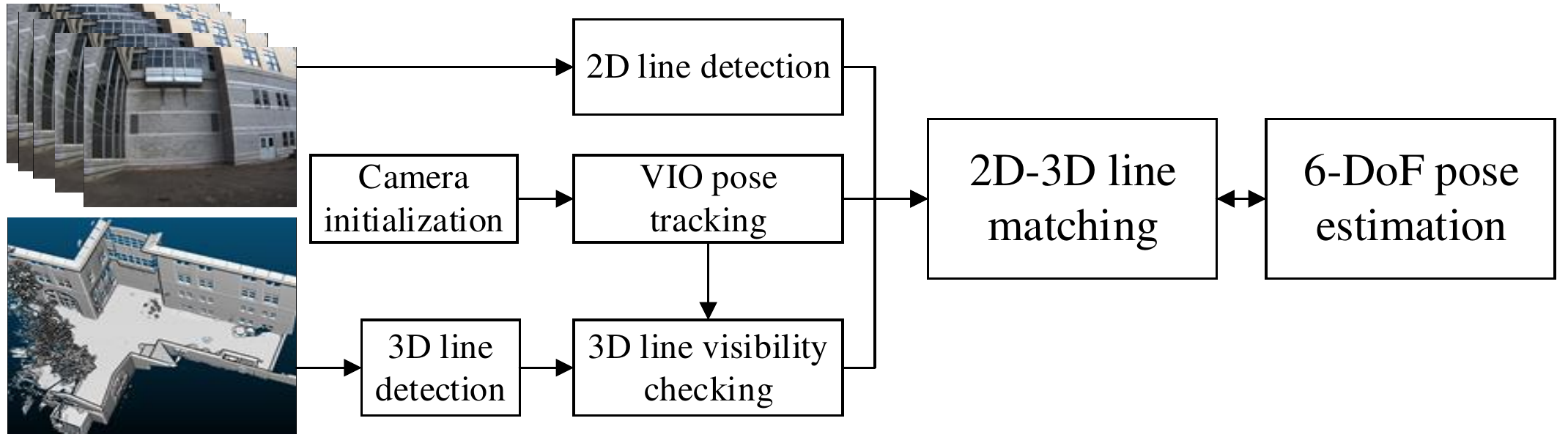}
	\end{center}
	\caption{Pipeline of the proposed camera localization method}
	\label{fig:overview}
\end{figure}

 The proposed method simultaneously estimates 6-DoF camera poses and 2D-3D line correspondences in LiDAR maps. The correspondences are utilized to optimize camera poses by minimizing the 3D line projection error while refined camera poses can help reject outlier correspondences. As the preliminary to the online 2D-3D correspondence estimation, 3D line features are extracted offline on the large scale 3D LiDAR maps. At the same time, a coarse pose initialization is given for the first frame by the PnP solver \cite{lepetit2009epnp} on manually labeled 2D-3D point correspondences. Then VINS-Mono \cite{qin2018vins} is utilized to predict the camera motion between adjacent keyframes. With the predicted poses, local 3D lines in camera field-of-view (FoV) are extracted and directly matched with the online extracted 2D lines from image sequences. Finally, camera poses and 2D-3D correspondences are iteratively updated. The pipeline is shown in Fig. \ref{fig:overview}. %84
 
\subsection{2D and 3D line extraction}
In urban environments, the geometric structures are often represented by line segments and planes. We use a segmentation-based 3D line detection method \cite{lu2019fast} to extract 3D lines from LiDAR maps. The general idea is to cluster point clouds into plane regions and use contour line fitting to obtain 3D line segments. This method is efficient and robust for large scale unorganized point clouds. Although it needs time to process millions of points, the 3D lines of all maps are extracted only once before we start tracking. %93

\begin{figure}[htp]
	\centering
	\begin{minipage}[b]{0.49\linewidth}
		\centering
		\includegraphics[width = .9\columnwidth]{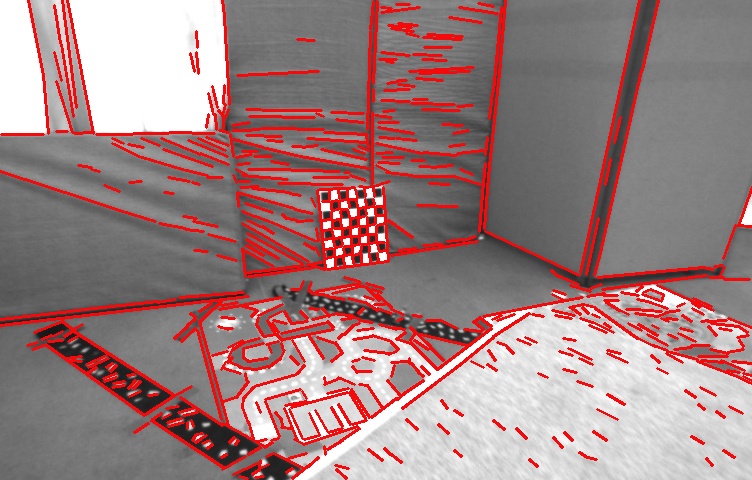} 
		\centerline{\scriptsize(\textbf{a}) Traditional LSD \cite{von2008lsd}} 
		\label{fig:lsd}
	\end{minipage}
	\begin{minipage}[b]{0.49\linewidth}
		\centering
		\includegraphics[width = .9\columnwidth]{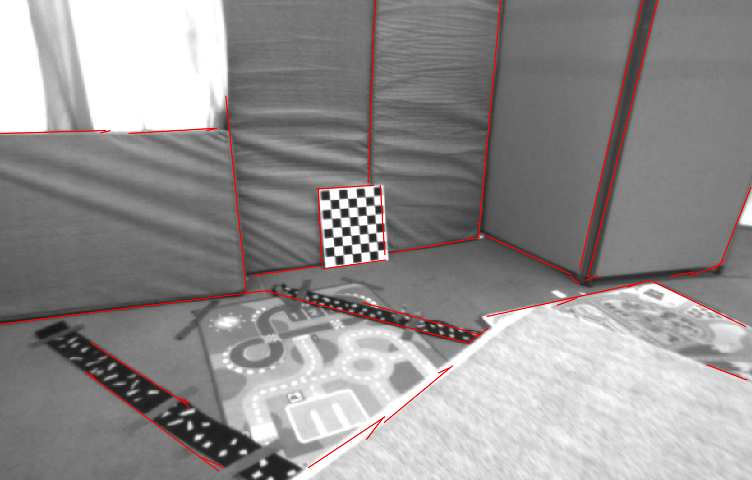}
		\centerline{\scriptsize(\textbf{b}) AFM line detection \cite{xue2019learning}} 
		\label{fig:afm}
	\end{minipage}
  
	\caption{Comparison of different methods for 2D line segment detection.}
	\label{fig:lineseg}
\end{figure}

For 2D line extraction, we want to extract the major geometric 2D lines which are consistent with 3D lines and robust to noises. It is challenging in urban scenes because the substantial texture noises yield fragmented 2D line segments and some geometric edges are invisible in 2D images on the color-homogeneous structures (e.g., white wall). Many state-of-the-art line segment detection (LSD) methods have been presented in computer vision \cite{von2008lsd, xue2019learning}, where traditional hand-craft methods are with high efficiency for online running on CPU. However, the detected lines are fragmented and noisy, an example is shown in Fig. \ref{fig:lineseg}(a). These kinds of fragmented and noisy features can produce a lot of 2D-3D matching outliers. Considering the line completeness and robustness to noisy, we finally employ a learning-based LSD \cite{xue2019learning} which uses the attraction field map (AFM) to transfer the LSD problem to be a region coloring problem. For each pixel $p$ in images, the model first learns a 2D vector ${\mathbf a}(p)$ from the pixel to its nearest point $p'$ on the nearest line segment. %81
\begin{equation}
    \mathbf{a}(p)=\boldsymbol{p}'-\boldsymbol{p}
\end{equation}
where $\boldsymbol{p}=(x,y)$ is the coordinate of pixel $p$. Then an AFM $\{\mathbf{a}(p), p\in I\}$ is generated by encoding the pixels to line associations. During the training step, ground-truth 2D lines are transferred to AFM representation and then the network directly learns a model to minimize the similarity between the output and the true AFM. 
% \begin{equation}
%     l(\mathbf{\hat{a}}, \mathbf{a})=\sum_{(x,y)\in I}||\mathbf{a}(p)-\mathbf{\hat{a}}(p)||_1
% \end{equation}
For testing, AFM representations are first generated from images, then a squeeze module \cite{von2008lsd} is used to iteratively group $\boldsymbol{v}(p)=\boldsymbol{p}+\mathbf{a}(p)$ belonging to the same line to fit line segments. Fig. \ref{fig:lineseg}(b) shows the detected line segments of the example image. It shows good consistency with geometric 3D structures and robustness to texture noises.

\subsection{2D-3D line matching}
\label{2D-3Dmatching}
For a single frame, the main steps to obtain 2D-3D correspondences consist of initial camera pose prediction, visible 3D lines collection, and individual 2D-3D line correspondence estimation. Here the extraction of 3D lines in FoV helps improve the efficiency because local 3D lines in FoV are very limited compared with all 3D lines in the 3D map. Considering the occlusion checking is difficult to conduct on only 3D lines map, we keep all the 3D lines in FoV without discarding occluded lines.  

For an image at time ${t}$, the corresponding pose estimation from VINS-Mono is denoted as $\bar{\mathbf{P}}_t\in SE(3)$, and the updated pose using 2D-3D correspondences is denoted as $\mathbf{P}_t \in SE(3)$. By using the estimated pose from 2D-3D correspondences of the last frame $\mathbf{P}_{t-1}$ and the camera motion from VINS-Mono $\mathbf{T}$, the updated pose $\mathbf{\hat{P}}_t$ can be computed
\begin{equation}
\begin{aligned}
    &\mathbf{T}=\bar{\mathbf{P}}_t\cdot (\bar{\mathbf{P}}_{t-1})^{-1}, \\
    &\hat{\mathbf{P}}_t=\mathbf{T}\cdot \mathbf{P}_{t-1}.
\end{aligned}
\end{equation}
With the pose prediction $\mathbf{\hat{P}}_t$, the local 3D lines $\{L_t\}$ in FoV can be extracted based on the two endpoint projections for efficiency. For checking the visibility of a 3D point $\boldsymbol{P}$ in FoV,
\begin{equation} \label{ptp}
    \boldsymbol{p}_t=\left[\begin{matrix}x&y&w\end{matrix} \right]^T= \mathbf{K}\cdot\mathbf{\hat{P}}_t\cdot\boldsymbol{P}
\end{equation}
where $\mathbf{K}$ is the camera intrinsic matrix. If $w>0$, $0\leq\lfloor x/w \rfloor< column $ and $0\leq\lfloor y/w\rfloor<row$, the 3D point is in FoV. For a 3D line segment, the visibility checking is more complicated. Based on the visibility of two endpoints, there are three cases for validation:
\begin{figure}[htbp]
	\begin{center}
		\includegraphics[width=0.90\linewidth]{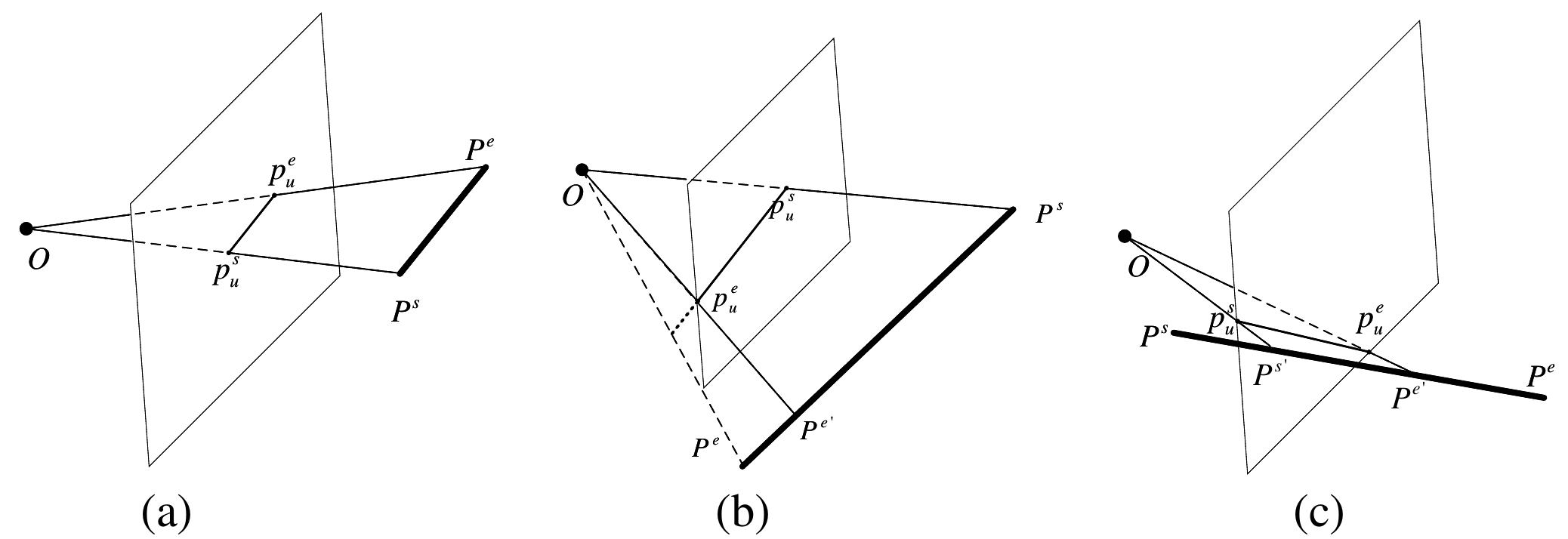}
	\end{center}
	\caption{3D line visibility checking}
	\label{fig:visible3d}
\end{figure}
\begin{itemize}
    \item If both two endpoints are in FoV (Fig. \ref{fig:visible3d}(a)), we keep the whole 3D line segment as a local visible feature.
    \item When only one endpoint is in FoV (in Fig. \ref{fig:visible3d}(b)), we iteratively sample new 3D points on the 3D line from the visible point by $0.1$ ratio of the line length and check the visibility of the new sample points. The generated subset 3D line segment with the longest length in FoV is stored as a local visible feature.
    \item When both two endpoints are out of FoV but a subset is in FoV, as shown in Fig. \ref{fig:visible3d}(c), we can also sample points to extract a subset. However, most of the invisible map lines are in this case, we discard all the 3D segments with two endpoints out of FoV for efficiency.
\end{itemize} %77

With the predicted camera pose $\mathbf{\hat{P}}_t$, 2D lines $\{l_t=(\boldsymbol{p}_t^s, \boldsymbol{p}_t^e)\}$ are directly matched with the local 3D lines in FoV, where $\boldsymbol{p}^s$ and $\boldsymbol{p}^e$ are the start and end points of a 2D line, respectively. For each possible 2D-3D correspondence, we use a 3D vector to measure the similarity, 2D angle distance $\theta$, the distance $d$ of two 3D endpoint projections to the corresponding infinite 2D lines, and the overlapped length $len_{overlap}$ of the finite 2D lines with the 3D line projection. 
\begin{equation}
    dist(L_i, l_j)=\{\theta, d, len_{overlap}\}
\end{equation}
The projection of a 3D line $L=(\boldsymbol{P}^s, \boldsymbol{P}^e)$ on the image plane is $l_u=(\boldsymbol{p}_u^s, \boldsymbol{p}_u^e)$ (Eq. \ref{ptp}), which is matched with the detected 2D line $l_t$. The normalized orientation of 2D lines is denoted as $\boldsymbol{v}=(\boldsymbol{p}^e-\boldsymbol{p}^s)/||\boldsymbol{p}^e-\boldsymbol{p}^s||$. Then the 2D angle distance $\theta$ can be computed by
\begin{equation}
    \theta =  \arccos (\boldsymbol{v}_t^T\cdot\boldsymbol{v}_u). 
\end{equation}
Assume the parametric representation of a extracted 2D line $l_t$ is $Ax+By+C=0$. The distance $d$ can be computed by
\begin{equation}
    d  = \frac{|Au^s_x+Bu^s_y+C|+|Au^e_x+Bu^e_y+C|}{\sqrt{A^2+B^2}}.
\end{equation}
By using the point-to-line projection points, the overlapped length $len_{overlap}$ between 3D line projections $l_u$ and the detected 2D lines $l_t$ is
\begin{equation}
\begin{aligned}
    &\alpha_* =\argmin_{\alpha\in[0,1]}||\boldsymbol{p}_t^s+\alpha\cdot(\boldsymbol{p}_t^e-\boldsymbol{p}_t^s)-\boldsymbol{p}_u||_2^2, \\
    &len_{overlap} =|\alpha_*^s-\alpha_*^e|\cdot||\boldsymbol{p}_t^e-\boldsymbol{p}_t^s||_2^2,
\end{aligned}
\end{equation}
where $\alpha_*^s$ and $\alpha_*^e$ are corresponding to the projection points of $\boldsymbol{p}_u^s$ and $\boldsymbol{p}_u^e$ respectively. Here $\alpha^*$ provides an unified representation of point to finite line distance. Then for each extracted 2D line $l_t^i$, brute-force searching strategy is used to find a 3D line whose distance $\theta<\theta_0$ and $d<d_0$. Thus we obtain a set of coarse 2D-3D line correspondences for further pose estimation.

\subsection{Pose optimization and correspondence refinement}
For a single frame, the camera pose can be optimized by minimizing the point-to-infinite-line distance of the two 3D endpoint projections to corresponding 2D line distance.  The Lie algebra of the estimated camera poses $\mathbf{P}_t$ are denoted as $\boldsymbol{\xi}_t$. The coefficient vector of the infinite 2D line can be $\mathbf{H}=\left[\begin{matrix}A&B&C\end{matrix} \right]$. The object function is to minimize the projection errors between all the 2D-3D correspondences:
\begin{equation} \label{costs}
\begin{aligned}
\boldsymbol{\xi}^{\star}_t&=\argmin_{\boldsymbol{\xi}_t}\sum_{i=1}^{M}d_i \\
&=\argmin_{\boldsymbol{\xi}_t} \frac{1}{2} \sum_{i=1}^{M}\frac{||\mathbf{H}_i\cdot \mathbf{K} \exp(\boldsymbol{\xi}_t) L_i||^2_2}{\sqrt{A_i^2+B_i^2}},
\end{aligned}
\end{equation}
where $L_i$ contains two endpoints, $M$ denotes the number of correspondences. It is formulated as a non-linear least-squares problem. We initialize the camera pose with the predicted pose $\mathbf{\hat{P}}_t$ from VINS-Mono. With Lie algebra, the typical L-M algorithm can find the optimal camera pose $\mathbf{P}_t$. 
% The projection of 3D points $\mathbf{K} \exp(\boldsymbol{\xi}) L_i$ is first transformed to inhomogeneous coordinates to compute point-to-infinite-line distance.

\begin{figure}[t]
	\begin{center}
		\includegraphics[width=0.99\linewidth]{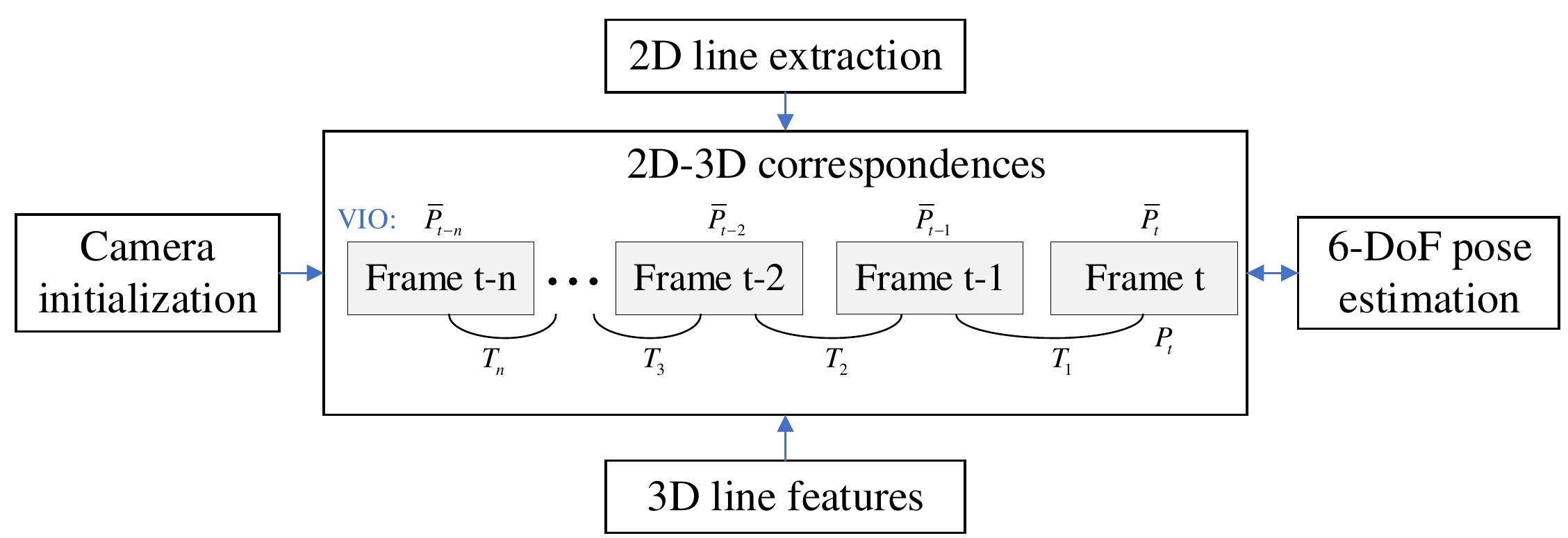}
	\end{center}
	\caption{Pose estimation in the sliding window}
	\label{fig:slidingwindow}
\end{figure}

% If one pose estimation is wrong, all the pose predictions for later frames will be far away from the right poses.
However, single frame 2D-3D correspondence observations are not robust enough for online camera localization. When the 3D lines in FoV are limited or parallel to each other in 3D space, the 2D-3D correspondences cannot constrain the 6-DoF pose. Additionally, even the correspondences are enough for pose estimation,  the geometric localization noises of both 2D and 3D lines will make the estimation jitter around the true pose. To solve these problems, a sliding window is utilized to add more previous correspondence observations for optimizing the current pose $\mathbf{P}_t$ (as shown in Fig. \ref{fig:slidingwindow}). Assuming the camera motion estimated from VINS-Mono for two adjacent keyframes is accurate, which is reasonable because the drifts of VINS-Mono on two adjacent keyframes are small. $\mathbf{T}_t^1, \mathbf{T}_t^2, \cdots, \mathbf{T}_t^n$ are denoted as camera motions between two adjacent keyframes ($4\times 4$ matrix including $R$ and $t$). The camera pose $\mathbf{P}_{t-n}$ of the previous $n$-th keyframe can be derived by the current estimation $\mathbf{P}_t$ and camera motion
\begin{equation}\label{eq::tf}
    \mathbf{P}_{t-n}=(\mathbf{T}_t^n)^{-1} \cdots (\mathbf{T}_t^2)^{-1} (\mathbf{T}_t^1)^{-1}\mathbf{P}_t.
\end{equation}

Then all previous 2D-3D correspondences in the sliding window can be ``visible" in current frame. In Eq. \ref{eq::tf}, the previous camera pose $\mathbf{P}_{t-n}$ is related to the current pose $\mathbf{P}_t$, while $(\mathbf{T}_t^1 \mathbf{T}_t^2 \cdots \mathbf{T}_t^{n} )^{-1}$ is a constant pose transformation, its Lie algebra is denoted as $\Delta \boldsymbol{\xi}_t^n$. Thus the pose optimization function is 
\begin{equation}
    \boldsymbol{\xi}^{\star}_t=\argmin_{\boldsymbol{\xi}_t} \frac{1}{2} \sum_{n=0}^{N}\sum_{i=1}^{M}\frac{||\mathbf{H}_i^n\cdot \mathbf{K} \exp(\boldsymbol{\xi}_t+\Delta \boldsymbol{\xi}_t^n) L_i^n||^2_2}{\sqrt{{A_i^n}^2+{B_i^n}^2}},
\end{equation}
where $N$ is the number of previous frames in the sliding window. Therefore, more observations make the estimation more robust to outliers, and the motion constraints smooth the pose trajectory. When using the sliding window, an unbalance problem will arise due to the different numbers of 2D-3D correspondences for each frame. To equalize the contribution of each frame correspondences and improve the efficiency, a threshold for the maximum number of correspondences is set to discard the 2D-3D correspondences with short overlap distance $len_{overlap}$. We employ Ceres Solver \cite{ceres-solver} to implement the optimization. After an optimized camera pose is obtained, the 2D-3D correspondences can be re-estimated by the updated pose to reject outliers. It follows the 2D-3D line matching in Section. \ref{2D-3Dmatching} with more restrict thresholds (e.g., $\theta_0=0.8*\theta, d_0=0.8*d_0$). Then a more accurate camera pose can be updated with the new correspondences. After several iterations, both camera pose and 2D-3D correspondences can be optimized.

\section{Experimental results}
The proposed method was tested on two different real-world dataset. The first experiment was conducted on the public available EuRoC MAV Dataset \cite{burri2016euroc} with ground-truth trajectories. Then, we performed real experiments on our dataset collected by a Realsense D435i camera under varying conditions to validate the performance.

\subsection{Implementation}
Considering the 3D LiDAR maps are often large scale with a big data volume, we first down-sample the point clouds using CloudCompare \cite{girardeau2011cloudcompare} with a constant space resolution (1-5cm in experiments). Then the 3D line extraction based on \cite{lu2019fast} is conducted to obtain 3D lines. Since the arbitrary pose initialization in maps is a kidnapped robot problem for global 2D-3D correspondence estimation, a coarse pose initialization is given for the first frame by the PnP solver \cite{lepetit2009epnp} on manually labeled 2D-3D point correspondences (at least 4 pairs, we use 6 pairs for robustness). For the learning-based 2D line extraction, we directly use the trained U-Net model  \cite{xue2019learning} from Wireframe dataset \cite{huang2018learning} and modify it for online line segment detection. The testing platform is a desktop with Intel Core i7-4790K CPU, 32GB RAM, and an Nvidia GeForce GTX 980Ti GPU. The GPU is only used for 2D line detection. %86

During the 2D-3D correspondence estimation, the angle distance threshold $\theta_0$ is set as 10 degrees to constrain the 3D projections to be almost parallel with corresponding 2D lines. Then the point to line distance threshold $d_0$ is set around $20\sim 30$ pixels to collect matching pairs. If the number of correspondences for an image is less than a threshold (set as 8 empirically), it is identified unstable therefore the camera pose is predicted by camera motion only. If the number exceeds a threshold, we discard the correspondences with short $len_{overlap}$ distance for efficiency  (empirically $M_0=40$, window size $N=10$, there will be at most $440$ correspondences in the sliding window). In the experiments, 2 or 3 iterations of optimization are often enough to obtain stable camera pose and 2D-3D correspondences. %82

Since our method is based on monocular visual-inertial odometry without loop closure, we compare it with the two versions of VINS-Mono \cite{qin2018vins}, i.e., with or without loop closure. Loop closure helps for reducing the overall absolute trajectory error and mapping, but the refinement of the past poses in the loop does not help for realtime localization. Additionally, it causes pose shifts for current pose which have side effects for robot navigation. 
% It should be noticed that it is unfair to compare a camera localization method with a VIO method. Camera localization methods use map information while VIO does not. Here we directly compare with the VINS-mono(odom) and VINS-mono(loop) to support the claims that out method does not have accumulated drifts or pose jumps. %86

%Currently, there is no other open sourced code of monocular camera localization in 3D LiDAR maps.  
\subsection{Result on EuRoC MAV Dataset}
\begin{figure}[htbp]
	\begin{center}
		\includegraphics[width=0.99\linewidth]{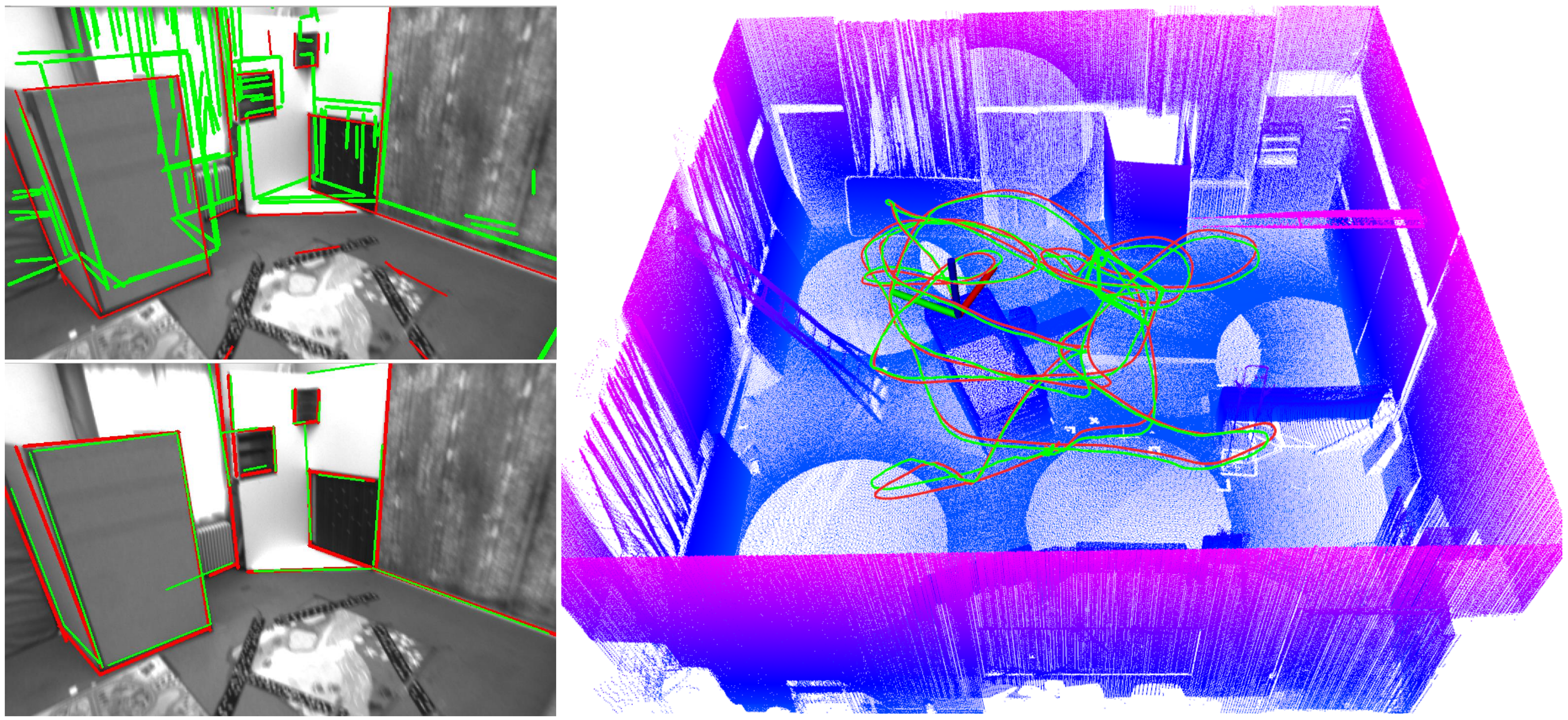}
	\end{center}
	\caption{Camera localization results on the EuRoC MAV dataset. The top left image shows the extracted 2D lines (red) and the projected 3D lines (green) using VINS-Mono(odom) (with occlusions), bottom left shows the 2D-3D correspondences using our method. The right image shows the estimated trajectory (green) aligned with ground-truth (red) in LiDAR maps.}
	\label{fig:euroc}
\end{figure}

The EuRoC MAV Dataset \cite{burri2016euroc} is a visual-inertial datasets collected onboard a UAV. The datasets contain stereo images (20Hz), synchronized IMU data (200Hz), accurate ground-truth trajectories about 70 meters and a scan LiDAR map. The subset room data consist of 2 LiDAR maps wherein 3 video sequences each. 2D-3D correspondence results and estimated trajectory in LiDAR maps are shown in Fig. \ref{fig:euroc}. All the 3D lines in FoV are projected to the image plane without occlusion checking. From the top left image, we can observe that the projections of 3D lines (in green) are shifted by using the estimated pose of VINS-Mono(odom). Then by iteratively updating 2D-3D correspondences and camera poses, stable correspondences are obtained with a more accurate pose. The position drifts are greatly reduced and stable 2D-3D correspondences can be estimated. 
% Since the ground-truth trajectory is not at the same frame with the LiDAR map, we align the first estimated camera pose with the ground-truth trajectory. %89
\begin{table}[htbp]
	\caption{{ATE RMSE \cite{sturm2012benchmark} results over 5 runs on the EuRoC MAV Dataset.}}
	\centering
	{
		\small
		\begin{tabular}{c|cccccc}
			\hline
        \multirow{2}{*}{Dataset} & VINS-Mono & VINS-Mono& 2D-3D\\
			                     & (odom)\cite{qin2018vins} & (loop)\cite{qin2018vins} & matching  \\
		\hline
		V1$\_01\_$easy  & 0.147  & \bf{0.075}  & 0.089  \\ 
		V1$\_02\_$medium& 0.153  & 0.127  & \bf{0.069} \\
		V1$\_03\_$difficult & 0.301 & \bf{0.152} & 0.173 \\
		V2$\_01\_$easy & 0.275 & \bf{0.099} & 0.166 \\
		V2$\_02\_$medium & 0.221 & 0.135 &  \bf{0.132} \\
		V2$\_03\_$difficult & 0.726 & \bf{0.324} & 0.635 \\
			\hline
	\end{tabular}}
	\label{tab:ate}
\end{table} 
% \begin{figure}[htbp]
% \setlength{\abovecaptionskip}{-0.3cm}   
% \setlength{\belowcaptionskip}{-1.0cm}
% 	\begin{center}
% 		\includegraphics[width=0.99\linewidth]{all_translation_rmse.pdf}
% 	\end{center}
% 	\caption{Boxplot of the ATE and variance over 5 runs on the EuRoC MAV Dataset }
% 	\label{fig:boxate}
% \end{figure}

For quantitative analysis, the beginning 200 estimated poses of each sequence are used for trajectory alignment with ground-truth \cite{zhang2018tutorial}. The absolute trajectory error (ATE) \cite{sturm2012benchmark} results are shown in Table \ref{tab:ate}, it is clear that the 2D-3D correspondences improve the pose estimation accuracy compared with only odometry. The V2 room has more noises making the 2D-3D correspondences sometimes unstable for pose optimization. This is the reason why the improvements for the V2 room are less significant than the results of the V1 room. The worst case is that no stable 2D-3D correspondence is available and the final estimations follow the odometry. Furthermore, our method shows competitive results compared with VINS-Mono(loop). While the loop closure optimizes the past poses in the loop and produces a pose jump for the current pose. For the realtime localization purpose, the refinements of the poses in the past make no sense and the pose jumps have side effects for navigation. Our method always estimates the current poses in the sliding window, which greatly reduces the drifts and does not have the issue of pose jumps. 

% It is reasonable because loop closures use all the observations in the loop to optimize the past poses while our method always optimizes the current poses. For the realtime localization purpose, the loop closure cannot change the robot poses in the past and also produce ``pose jumps" for current poses which has bad effects for navigation. Our method always optimizes the current pose in the sliding window, which greatly reduces the drifts and does not have the issue of ``pose jumps". 

\begin{figure}[htbp]
	\begin{center}
		\includegraphics[width=0.99\linewidth]{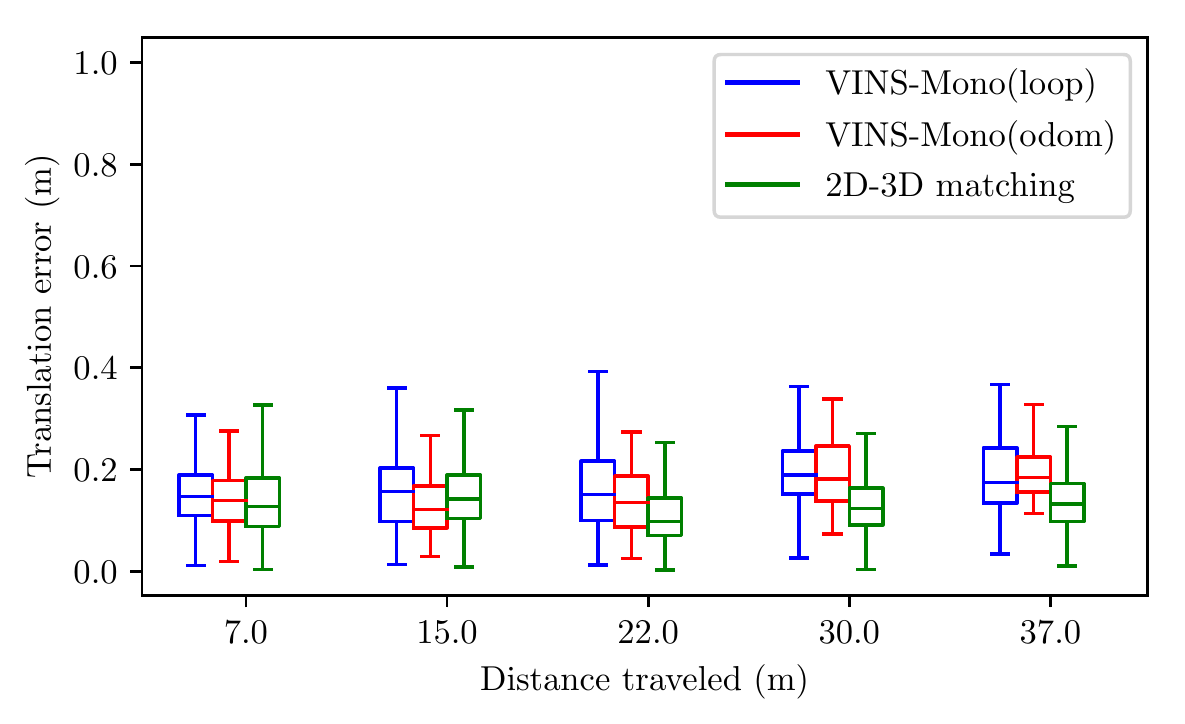}
	\end{center}
	\caption{Boxplot of the relative trajectory errors}
	\label{fig:rpe}
\end{figure}

The average relative pose errors (RPE) \cite{zhang2018tutorial} are shown in Table \ref{tab:rpe} and Fig. \ref{fig:rpe}, which are used to show the growth of position error with the trajectory length. The accumulated drift of VINS-mono(odom) is growing up with travel length. While the error of our method keeps small and stable along the way, which is related to the accuracy of 2D and 3D line localization.
\begin{table}[htbp]
	\caption{RPE RMSE \cite{sturm2012benchmark} over different segment lengths}
	\centering
	{
		\small
		\begin{tabular}{c|cccccc}
			\hline
        Segment & VINS-Mono & VINS-Mono& 2D-3D\\
		Length(m)  & (odom)\cite{qin2018vins} & (loop)\cite{qin2018vins} & matching  \\
		\hline
		 7  & \bf{0.150}  & 0.166  & 0.173  \\ 
		15& 0.161  & \bf{0.154}  & 0.172 \\
		22 & 0.184 & 0.174 & \bf{0.139} \\
		30 & 0.200 & 0.1840 & \bf{0.158} \\
		37 & 0.212 & 0.190 & \bf{0.168} \\
			\hline
	\end{tabular}}
	\label{tab:rpe}
\end{table} 
%  The rolling shutter RGB camera is not used because the VINS-Mono does not work well on this camera.
\subsection{Evaluation on our dataset}
\begin{figure}[htbp]
	\begin{minipage}[b]{0.32\linewidth}
		\centering
		\includegraphics[width = .9\columnwidth]{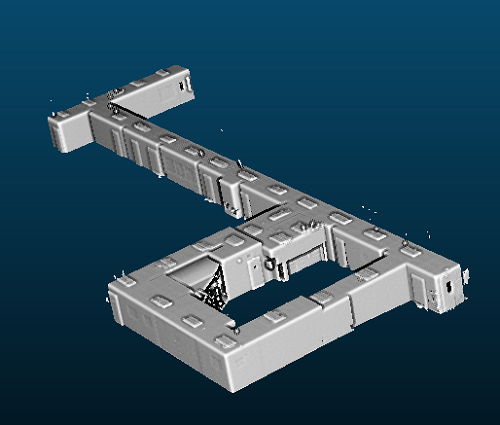} 
		\centerline{\scriptsize(\textbf{a}) Corridors} 
		\label{fig:corridors}
	\end{minipage}
	\begin{minipage}[b]{0.32\linewidth}
		\centering
		\includegraphics[width = .9\columnwidth]{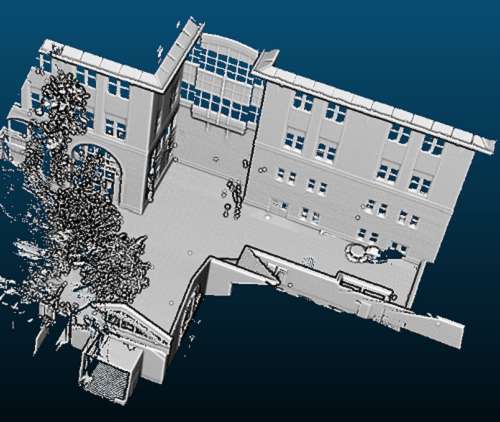}
		\centerline{\scriptsize(\textbf{b}) NSH building} 
		\label{fig:nshwall}
	\end{minipage}
	\begin{minipage}[b]{0.32\linewidth}
		\centering
		\includegraphics[width = .9\columnwidth]{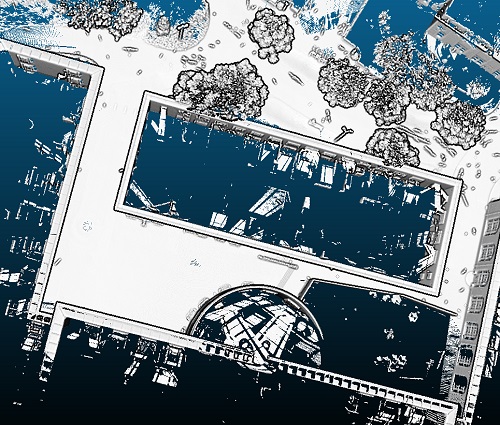}
		\centerline{\scriptsize(\textbf{c}) Smith Hall} 
		\label{fig:smith}
	\end{minipage}
	\caption{Three urban scene LiDAR maps.}
	\label{fig:map}
\end{figure}

 To further evaluate our method under various environments, we tested it on our own collected data of indoor corridors and outdoor buildings. An Intel RealSense D435i camera is used to collect synchronized images and IMU data. The left global-shutter imager captures monocular image sequences ($640\times 480 $ pixels images at 30Hz, the IR projector turned off) with synchronized IMU data (200Hz). The LiDAR maps are obtained by registering several scans of a FARO scanner focus3D S, as shown in Fig. \ref{fig:map}. Both the indoor corridors (Fig. \ref{fig:map}a) and Smith Hall (Fig. \ref{fig:map}c) have a lot of occlusions, while the NSH building (Fig. \ref{fig:map}b) is much simpler with fewer occlusions. For these experiments, the trajectories are in the same pattern to run a complete round and return to the start points to see the position drifts. The loop closure is not stably detected for all runs, so the results of VINS-Mono with loop closure are not discussed.
\begin{figure}[htbp]
	\begin{center}
		\includegraphics[width=0.98\linewidth]{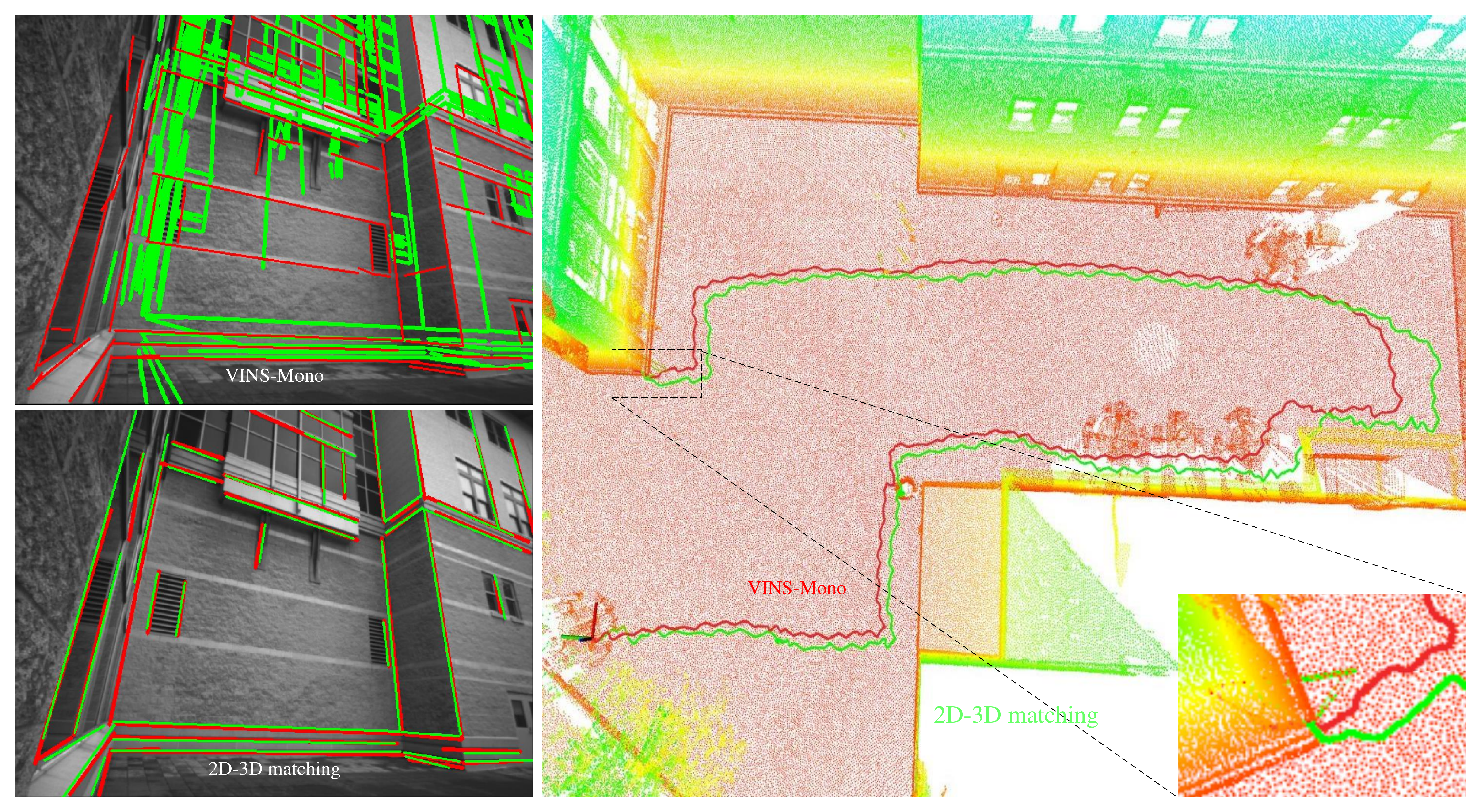}\\
		\vspace{0.1cm}
		\includegraphics[width=0.98\linewidth]{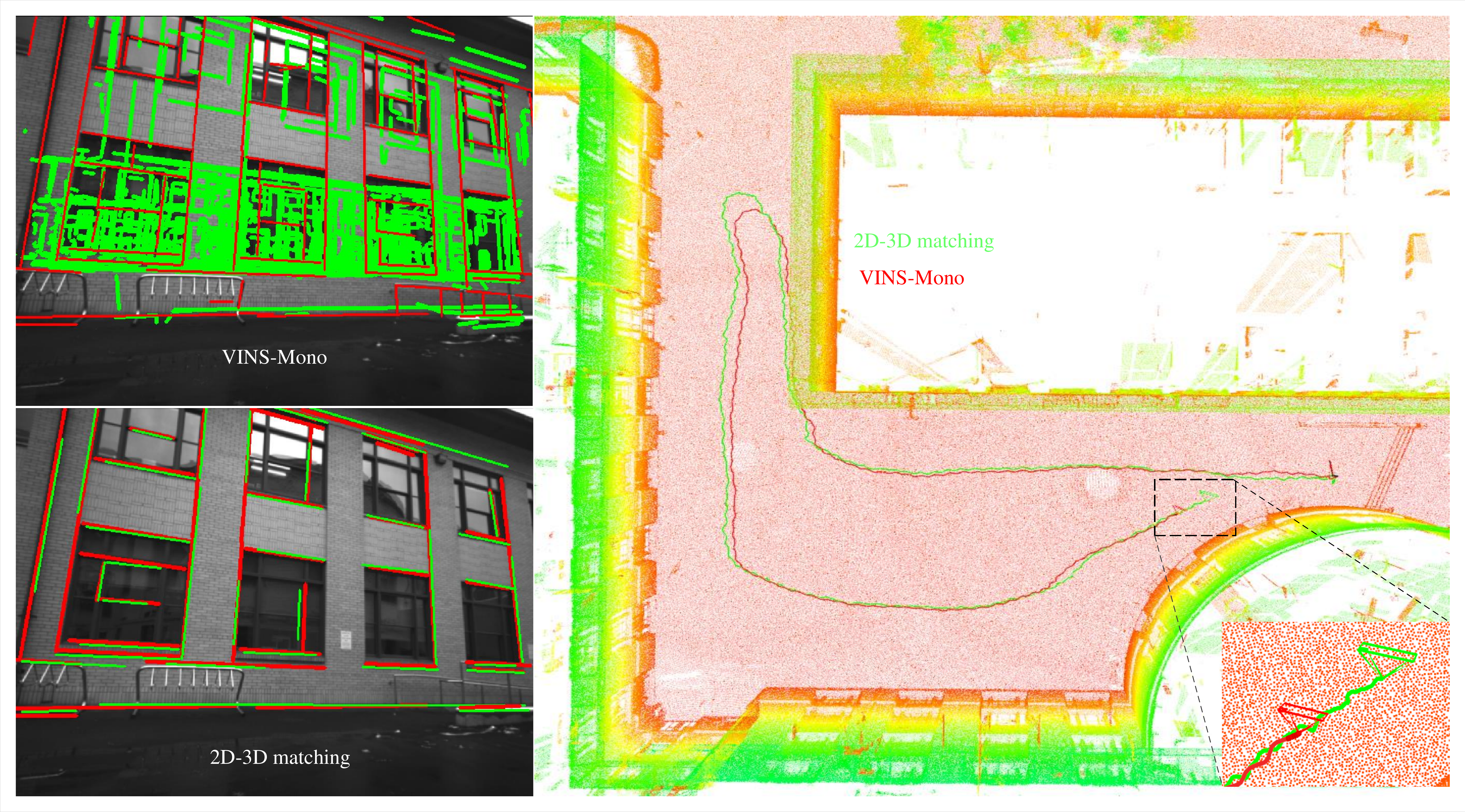}
	\end{center}
	\caption{Camera localization results in outdoor environments}
	\label{fig:nsh}
\end{figure}
%  It should be noticed that both VINS-Mono and our method start from the same position and the same pose initialization, thus both methods show good 2D-3D line structure alignments at the beginning. the visualization of 3D line projections (green lines) and the 2D lines (red lines) are shown on the top left images. I

 The indoor corridor's results are shown in Fig. \ref{fig:corridor}. The results of two outdoor buildings are shown in Fig. \ref{fig:nsh}. Considering that we do not have ground-truth trajectories, the estimation accuracy validation is shown in the following two ways. For the qualitative analysis, the 3D line features are projected to overlap with 2D lines using the estimated poses of VINS-Mono(odom) and our method. For VINS-Mono odometry, it can be observed that the 3D line projections (green lines) are shifted and scaled by inaccurate camera poses in the top left images. While using our method, the pose estimations provide more accurate and stable 2D-3D structure correspondences in the bottom left images. Additionally, we can observe obvious misalignment between the two trajectories. More online correspondence and localization results are shown in the video \footnote{\href{https://youtu.be/H80Bnxm8IPE}{https://youtu.be/H80Bnxm8IPE}}. 
 
 For the accuracy evaluation, we pick 5 frames along the trajectories and use the PnP solver to estimate the ground-truth camera poses on manually labelled 10 pairs of 2D-3D point correspondences. The 5 frames are sampled along the trajectories on different times ($t=\beta*t_0$, $t_0$ is the total running time) and the RMSE positions along 5 runs are used as the final ground-truths. The position errors are shown in Table \ref{tab:rs}. The accumulated drifts increase a lot along the trajectories for VINS-Mono(odom). While our method greatly improves the localization accuracy with the assistance of stable 2D-3D line correspondences. The localization errors keep small along the whole trajectories. Another interesting observation is that the accumulated error can drift back if we keep the direction of the system and backward to the start position, which is shown in the results of VINS-Mono on NSH building. %74 iteratively updates the 2D-3D correspondences and camera pose based on camera motion, which
 \begin{table}[htbp]
	\caption{Localization errors in man-made environments}
	\centering
	\setlength{\tabcolsep}{1.2mm}
	{
		\begin{tabular}{c|cc|cc|cc}
			\hline
		    \multirow{2}{*}{$\beta$} & \multicolumn{2}{c}{\text{Corridors}} & \multicolumn{2}{|c}{\text{NSH building}}& \multicolumn{2}{|c}{\text{Smith Hall}} \\
			  & VINS- & 2D-3D & VINS- & 2D-3D &VINS- & 2D-3D  \\ 
			 ($t=\beta t_0$) & Mono & matching &Mono & matching & Mono & matching \\
			  \hline
			0.1  & 0.125   & \bf{0.109} & \bf{0.137} & 0.152 & 0.354 & \bf{0.180}    \\ 
			0.3 & 0.596 & \bf{0.128} & 0.319 & \bf{0.127} & 0.507 & \bf{0.154}     \\
			0.5   & 0.438 & \bf{0.094} & 1.203 & \bf{0.170} & 1.057 & \bf{0.201}   \\
			0.8 &  0.575 & \bf{0.120} & 0.352 & \bf{0.156} & 1.245 & \bf{0.156}  \\
			1.0 &  0.705 & \bf{0.112} & 0.621 & \bf{0.132} & 2.176 & \bf{0.178} \\
			\hline
			Length(m) &\multicolumn{2}{c|}{130}&\multicolumn{2}{c|}{95}& \multicolumn{2}{c}{120}\\
			\hline
	\end{tabular}
	}
	\label{tab:rs}
\end{table} 
 
In terms of efficiency, VINS-Mono does not use map information and can be customized set for output frequency. However, with different setting frequencies, the odometry results change a lot. We select the most stable one at 15 Hz. Then for the estimation of 2D-3D correspondences and the camera poses, it costs about 0.01 seconds on average for each keyframe. Since the 3D line extraction is offline before the system starts, 2D line detection can run at 25Hz on $640\times 480 $ images, our method can run at about $13\sim15$ Hz for all the scenarios.

\section{Conclusion}
\label{sec5}
In this paper, we presented a novel monocular camera localization approach in prior LiDAR maps of structured environments. With the 3D geometric lines from LiDAR maps and robust online 2D line detection, our method efficiently obtains coarse 2D-3D line correspondences based on the camera motion prediction from VINS-Mono. The pose optimization with 2D-3D correspondences greatly reduces the pose estimation drifts of VIO system without using visual-revisiting loop closure. Both qualitative and quantitative results on real-world datasets demonstrate that our method can efficiently obtain reliable 2D-3D correspondences and accurate camera pose in LiDAR maps. As future work we intend to enhance the robustness of 2D-3D correspondences on inaccurate pose predictions, such as directly using the pose of the last frame as the prediction.

%  With the 3D geometric lines from LiDAR maps and robust online 2D line detection, our method efficiently obtains coarse 2D-3D line correspondences based on the camera motion prediction from VINS-Mono. The camera poses and 2D-3D correspondences are iteratively refined by minimizing the projection errors of all 2D-3D correspondences in the sliding window.

% In the future, one direction is to detect stable and accurate 3D geometric features from noisy reconstructed point cloud maps using visual or LiDAR SLAM. Another direction is to

%%%%%%%%%%%%%%%%%%%%%%%%%%%%%%%%%%%%%%%%%%%%%%%%%%%%%%%%%%%%%%%%%%%%%%%%%%%%%%%%

\section*{ACKNOWLEDGMENT}
This work is supported by the Shimizu Institute of Technology, Tokyo and China Scholarship Council.% The authors also want to thank Warren Whittaker for the instructions on using FARO scanner.

% {\color{black}Sincere thanks are given to the anonymous reviewers and members of the editorial team for their comments and valuable recommendations. }
% The authors also want to thank Warren Whittaker from CMU for the instructions on using FARO scanner and Dylan Campbell from ANU for the discussions and helps. %Huai Yu is supported by China Scholarship Council. 
%%%%%%%%%%%%%%%%%%%%%%%%%%%%%%%%%%%%%%%%%%%%%%%%%%%%%%%%%%%%%%%%%%%%%%%%%%%%%%%%
\bibliographystyle{IEEEtran}
\bibliography{IEEEabrv, egbib}
\end{document}